\documentclass{article}

\usepackage{microtype}
\usepackage{graphicx}
\usepackage{subfigure}
\usepackage{booktabs} 

\usepackage{hyperref}

\def\pos{^\texttt{+}}
\def\neg{^\texttt{-}}

 \usepackage[accepted]{icml2024}

\usepackage{amsmath}
\usepackage{amssymb}
\usepackage{mathtools}
\usepackage{amsthm}
\usepackage{caption}

\usepackage{listings}
\definecolor{codegreen}{rgb}{0,0.6,0}
\definecolor{codegray}{rgb}{0.5,0.5,0.5}
\definecolor{codepurple}{rgb}{0.58,0,0.82}
\definecolor{backcolour}{rgb}{0.95,0.95,0.92}

\lstdefinestyle{mystyle}{
    backgroundcolor=\color{backcolour},   
    commentstyle=\color{codegreen},
    keywordstyle=\color{magenta},
    numberstyle=\tiny\color{codegray},
    stringstyle=\color{codepurple},
    basicstyle=\ttfamily\footnotesize,
    breakatwhitespace=false,         
    breaklines=true,                 
    captionpos=b,                    
    keepspaces=true,                 
    numbers=left,                    
    numbersep=5pt,                  
    showspaces=false,                
    showstringspaces=false,
    showtabs=false,                  
    tabsize=2
}

\usepackage[capitalize,noabbrev]{cleveref}

\theoremstyle{plain}

\theoremstyle{definition}

\theoremstyle{remark}

\usepackage[textsize=tiny]{todonotes}

\icmltitlerunning{Preference Optimization with the Pairwise Cringe Loss}

\begin{document}

\twocolumn[
\icmltitle{{\em Some things are more {\sc Cringe} than others:}\texorpdfstring{\\}{}
Iterative Preference Optimization with the Pairwise Cringe Loss}




\begin{icmlauthorlist}
\icmlauthor{Jing Xu}{yyy}
\icmlauthor{Andrew Lee}{yyy}
\icmlauthor{Sainbayar Sukhbaatar}{yyy}
\icmlauthor{Jason Weston}{yyy}
\end{icmlauthorlist}

\icmlaffiliation{yyy}{Meta}

\icmlcorrespondingauthor{Jing Xu}{jingxu23@meta.com}

\icmlkeywords{Machine Learning, ICML}

\vskip 0.3in
]



\printAffiliationsAndNotice{}  

\begin{abstract}

 Practitioners commonly align large language  models using pairwise  preferences, i.e., given labels of the type response A is preferred to response B for a given input. 
 Perhaps less commonly, methods have also been developed for binary feedback,
 i.e. training models given labels of type response A is good or bad. We show how an existing performant binary feedback method, the Cringe Loss \cite{adolphs2022cringe}, can be generalized
 to the pairwise preference setting using a simple soft margin extension. 
Pairwise Cringe Loss is straightforward to 
implement and efficient to train, and we find it outperforms  state-of-the-art preference optimization algorithms such as PPO and DPO on the AlpacaFarm benchmark. 
We show that iterations of training of our model are important for improved results, and  that we can generalize DPO to Iterative DPO in the same way. 
\end{abstract}

\section{Introduction}

Aligning large language models (LLMs) after pre-training can give large gains in 
their performance for downstream tasks for users \cite{roller2020recipes,gururangan2020don,ouyang2022training}.
Exactly how to implement this alignment depends on the labels one collects.
Given positive examples of correct behavior one can perform 
supervised fine-tuning (SFT) using standard likelihood-based training.
Given both positive and negative examples (\textbf{binary} feedback), 
one can use methods such as unlikelihood training on the negative examples \cite{unlikelihood_training}, or
 the more performant Cringe Loss \cite{adolphs2022cringe}.
However, a more common approach than using binary feedback,
popularized by work such as 
\citet{stiennon2020learning,ouyang2022training,touvron2023llama} is to collect \textbf{pairwise} preferences of the type response A is better than response B for a given input. In this case one can use methods such 
as PPO \cite{schulman2017proximal}, DPO \cite{rafailov2023direct}
and other variants.

In this work we seek to compare SFT, 
binary feedback and pairwise preference algorithms, and to ask the question: can one convert existing binary feedback algorithms to use pairwise preference data?
In particular the Cringe Loss is a method for binary feedback, which we 
show can be generalized to the pairwise preference case.
The Cringe Loss works as follows:  positive examples use the standard likelihood training loss, while for a given negative example it contrasts each token in the negative sequence against other likely tokens -- to encourage the negative sequence to no longer be the top-ranked sequence. 
%
%
After training on the initial feedback data,
the method is  then iterated by labeling data using the improved model, which was shown to  improve results further. Cringe Loss was shown to perform well with binary feedback data compared to competing methods, such as SFT, unlikelihood loss and best-of-N reranking
\cite{adolphs2022cringe} and for improving large-scale dialogue systems \cite{xu2023improving}.  However, collecting and using pairwise preferences for training has currently proven a more popular approach to developing aligned LLMs.

We thus explore generalizing the Cringe Loss to the pairwise preference setting.
We hence develop the Pairwise Cringe Loss,  by using a differentiable margin-based loss on the pair of responses.
In particular, we add a margin-based multiplier to the Cringe Loss to turn it on or off  depending on how much probability distance is between the pair.
When the preferred response A becomes much more likely than the response B, the Cringe Loss is turned off so that the model capacity is better spent on pairs that are closer in probabilities.
We can do multiple iterations of the Pairwise Cringe Loss training by generating new responses from the improved model and labeling them using a reward model if we have access to one.
A natural question is whether we can apply the same technique to methods like DPO as well. We hence also propose Iterative DPO that works in a similar manner, and compare to it in our experiments.

\begin{figure*}[t!]
    \centering
    \includegraphics[width=1\textwidth,trim={0 0.75cm 0 0}]{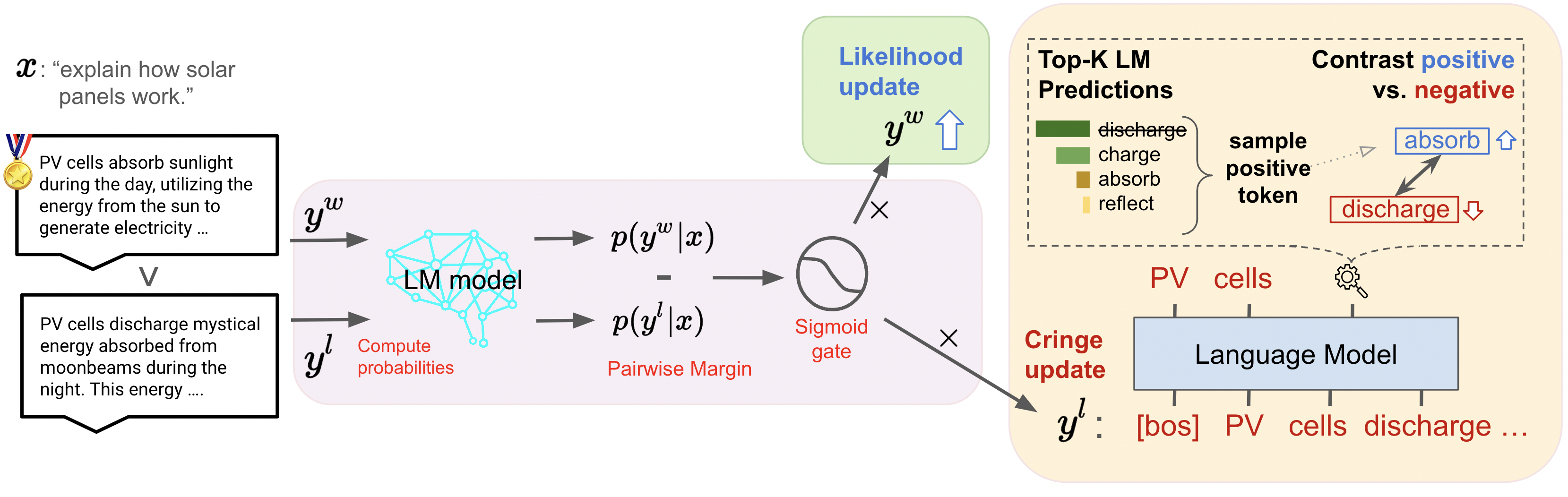}
     \vspace{1mm}
    \caption{\textbf{Pairwise Cringe Loss} update. We are given a preference pair of two documents: $y^w$,  preferred over $y^l$, for a given input $x$. The model likelihood of generating those responses $p(y^w|x)$ and $p(y^l|x)$ are used to form the pairwise margin in  \autoref{eq:margin}. A sigmoid is then used to weight the update of the pair, with a likelihood update being applied to $y^w$, and a cringe update to  $y^l$, see \autoref{eq:pair_loss}. The cringe update  penalizes the     output sequence of negative examples. For each negative token, a positive prediction is sampled from the language model to contrast against it.    }
    \label{fig:update_rule}
\end{figure*}

We experimentally compare competing approaches, including binary and pairwise variants of Cringe Loss. 
The first task is to
reduce repetitions \citep{arora2022director, unlikelihood_training},
which can be measured accurately so it gives us more control.
In this task, we find that Pairwise Cringe outperforms Binary Cringe,  and has performance similar to DPO, while
the Pairwise Cringe generations have slightly better quality.
Next, we employ a more realistic setup using the AlpacaFarm~\cite{dubois2023alpacafarm} benchmark that provides pairwise preference data for general instruction following.
Pairwise Cringe Loss again outperforms the Binary Cringe variant, in addition to SFT, and more importantly outperforms state-of-the-art methods  DPO and PPO, as well as the newly proposed Iterative DPO. 
Pairwise Cringe Loss is simple to implement and efficient to train, 
and is therefore a strong candidate for training instruction tuning and other alignment tasks.

\section{Preference Learning with the Cringe Loss}
\label{sec:method}

We first review the binary feedback-based (standard) Cringe Loss, and then 
introduce its generalization to the pairwise preference learning case.

\subsection{Standard Cringe Loss}\label{sec:binary_cringe}

The Cringe (ContRastive Iterative Negative GEneration) 
Loss is an approach developed 
for the binary feedback learning case, given
two sets of sequences: positive sequences $y\pos$, and negative sequences $y\neg$.
It is common for them to be responses to specific input sequences: $x\pos \rightarrow y\pos, x\neg \rightarrow y\neg$, i.e., given prompts or instructions $x$.
Note that the positive and negative labels only apply to the response portions.

The optimization objective consists of two terms: the cross entropy loss for the positive sequences and the Cringe Loss for the negative sequences. The former is used as standard, i.e., for all tokens $y_t\pos$ from a positive sequence $y\pos$:

\begin{align}\label{eq:cross_entropy}
\mathcal{L}_\text{CE}(x\pos,y\pos)  &= -\log p( [x\pos, y\pos] ) \\
&= -\log p(x\pos) - \log p(y\pos | x\pos) .
\end{align}
This will increase the likelihood of generating the positive responses.
Note that the loss included input tokens $x\pos$, but we can choose to only train on (update) the response portion $y\pos$ as well.

For a given negative sequence $y\neg$, the Cringe Loss contrasts each negative token $y_t\neg$ in the sequence against a positive token. It was argued in \citet{simple_contrastive_loss} that methods such as Unlikelihood \cite{unlikelihood_training} which simply push down the probability of negative tokens may  inadvertently push up the probability of low quality or rare tokens for that sequence position, because  there is no control over that effect.  
The Cringe Loss controls for this with its contrastive loss which 
instead encourages an alternative highly likely token to replace a given penalized token. However, in the training data one is typically provided a negative sequence, but one does not know for any given negative token in the sequence what an alternative positive token should be.

\begin{figure*}[t!]
    \centering
    \includegraphics[width=1\textwidth,trim={0 0.75cm 0 0}]{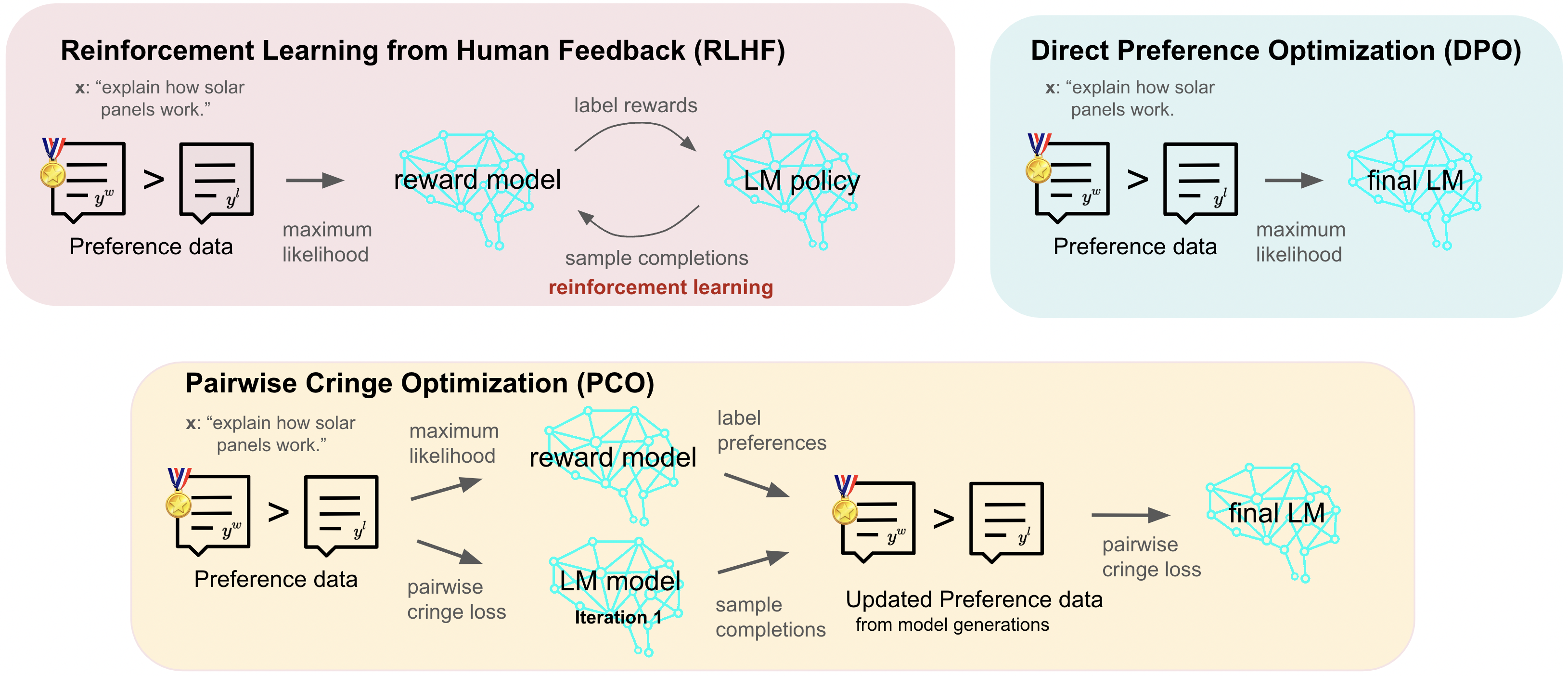}
    \vspace{1mm}
    \caption{\textbf{Pairwise Cringe Optimization (PCO)}. RLHF uses a reward model to label samples from the LM policy model as it trains. DPO optimizes for human preferences while avoiding reinforcement learning or a reward model. In contrast, PCO first directly optimizes using the original preferences to build an initial LM model, and then labels completions from that model with a reward model to build an updated preference dataset. This updated dataset is then used to train the final model using the Pairwise Cringe Loss.}
    \label{fig:pco}
\end{figure*}

The Cringe Loss thus proposes to sample an alternative positive token from the model's current top-k predictions (omitting the negative token, if it is in the top-k so that the same negative token is not chosen as the positive example).  
Let $s_t[i]$ be the model output score (input to the final softmax) at time $t$ corresponding to token $i$.
First we select top-k scores $\{s_t^1, ..., s_t^k\}$ from all scores $s_t[i]$ excluding the negative token $s_t[y_t\neg]$. 
Next we sample according to the categorical distribution constructed through the softmax over these top-k scores
\begin{equation}
s_t^* \sim \text{Softmax}(s_t^1, ..., s_t^k) .
\end{equation}
Now we can use $s_t^*$ as an alternative positive token.
The contrastive loss is then:
\begin{align}\label{eq:cringe}
\mathcal{L}_\text{Cr}(x\neg,y\neg)  &=- \sum_t \log \frac{\exp(s_t^*)}{ \exp(s_t^*) +  \exp(s_t[{y\neg_t}])},
\end{align}
which pushes down the negative token score to be smaller than the selected positive token.
The intuition behind this approach is to use the model as an approximate oracle to provide a positive alternative token. Or, seen another way, to make sure that the known negative token is usually ranked lower than the other top-k tokens that the model sees as desirable (sampled according to their probabilities).
This process can be seen in the right portion of \autoref{fig:update_rule} where negative token ``discharge'' is contrasted against a sampled positive token ``absorb''.


The final standard (binary feedback) Cringe Loss objective function for a single iteration is thus: 
\begin{align}
\mathcal{L}_\text{Bin}(x\pos, y\pos, x\neg, y\neg)  &= \mathcal{L}_\text{CE}(x\pos, y\pos) + \alpha \mathcal{L}_\text{Cr}(x\neg, y\neg)
\label{eq:binary_loss}
\end{align}
where $\alpha$ is a tunable hyper-parameter that controls the impact of the negative examples.

\paragraph{Iterative Training}
The negative sequences used for training either come from (i) human annotations, or (ii) access to a classifier (e.g., trained from
the human annotations), which can hence be seen as a reward model.
The latter can be used to iteratively label the model’s own generations and apply the Cringe Loss to those examples as well.
\citet{adolphs2022cringe} and \citet{xu2023improving} showed these iterations improve the models further.

\subsection{Pairwise Cringe Loss}

When given pairwise preference data, we are provided with samples
of the form $(x \rightarrow y^w, y^l)$, where the ``winning'' sequence $y^w$ has been labeled as {\em preferred} compared to the ``losing'' sequence  $y^l$ for the same input $x$.
For example, in instruction tuning tasks, such data is typically presented as two responses to the same instruction $x$, where one is preferred to the other as more helpful and/or harmless.

Let us define a margin between the two responses as
\begin{equation} \label{eq:margin}
    M(x, y^w, y^l) = \log p(y^w|x) - \log p(y^l|x).
\end{equation}
A negative margin means that the model is more likely to generate the losing sequence than the winning sequence, which is undesirable. In that case, we can employ the binary Cringe Loss from \autoref{eq:binary_loss} to push down the loser sequence probability while pushing up the winner sequence.
In contrast, when the margin is sufficiently large, the losing sequence is much less likely, so it becomes less important to push them apart anymore.
Therefore, we construct a loss that applies the binary Cringe Loss only when the margin is small using a sigmoid gate:
\begin{equation}
    g(x, y^w, y^l) = \sigma\left([b - M(x, y^w, y^l)] / \tau \right) 
\end{equation}
\begin{equation}
    \mathcal{L}_\text{Pair}(x,y^w, y^l) = g(x, y^w, y^l)  \mathcal{L}_\text{Bin}(x,y^w,x,y^l).
    \label{eq:pair_loss}
\end{equation}
Here, the gating function $g$ uses sigmoid  $\sigma$  to smoothly switch off the binary Cringe Loss for larger margins.
Its temperature $\tau$ controls the smoothness of this transition, while the bias $b$ determines how large a margin needs to be for the binary Cringe Loss to switch off. For example, a small $b$ value means the gating will turn off  even for small margins, thus the loss will be less aggressive in pushing the pairs apart.
In our experiments, we will  also compare it against a hard step function (a so called ``hard margin'', rather than a ``soft margin'').

Note that the gradient from the loss in \autoref{eq:pair_loss} has two pathways. The first one goes through the sigmoid multiplier and will act to increase the margin, which only depends on the sequence-level probabilities. The second gradient pathway is through the binary Cringe Loss and operates on token-level probabilities.
Therefore, this loss can be viewed as combining elements of  methods like DPO and PPO that operate only on sequence-level probabilities, and methods like Cringe and Unlikelihood that manipulate token-level probabilities -- while extending those latter methods to the pairwise preference case.

We note that we did not add a KL regularization term to our training objective, as is used in several other methods \cite{schulman2017proximal,rafailov2023direct} -- as we found experimentally our method already performs  well, and did not display issues of degradation without this term. However, it is possible  in certain settings adding such a term could improve performance, and hence could be considered.

We give an overall summary of the loss in \autoref{fig:update_rule}.
Code for implementing the Pairwise Cringe Loss is given in \autoref{sec:code}.

\paragraph{Iterative Training}
Like DPO, Pairwise Cringe Loss can be trained without a reward model given pairwise preference data using the recipe described above, which is the first iteration of Cringe training. However, like binary Cringe Loss, 
we can employ Pairwise Cringe Loss to perform iterative training. 
Our overall training approach, which we call Pairwise Cringe Optimization (PCO), is summarized in \autoref{fig:pco}.
Given a reward model that predicts preferences, the method is applied in an iterative manner:
\begin{enumerate}
\setlength{\itemsep}{0pt}
\item Train with the Pairwise Cringe Loss on the original preference data.
\item Generate new responses with the newly trained model (multiple responses per prompt $x$).
\item Label those responses  with the reward model, and choose new preference pairs. 
\item Train with the Pairwise Cringe Loss on a combination of the original preference data and the newly labeled data.
\end{enumerate}
Steps 2-4 can be repeated multiple times, however in our experiments in this paper we only perform these 4 steps (which we call 2 iterations).

To construct pairs we generate $N=4$ responses per input, and then choose the best and worst scoring as a single pair
using a scalar reward model (that assigns scores individually per response), discarding the other generated responses. However, other methods for assigning pairs are certainly possible that we have not explored. 

\section{Experiments}
\label{sec:experiments}

We first conduct experiments in \autoref{sec:repeat} on a repetition mitigation task from  \citet{arora2022director} in order to compare Pairwise Cringe Loss to the original Cringe Loss, as well  as comparing to DPO and other methods.
We then compare against  preference optimization methods for general instruction tuning, including comparing to PPO and DPO, on the AlpacaFarm benchmark in \autoref{sec:alpacafarm}.

\begin{figure}[t!]
    \centering
    \includegraphics[width=0.52\textwidth,trim={9mm 0 0 5mm}, clip]{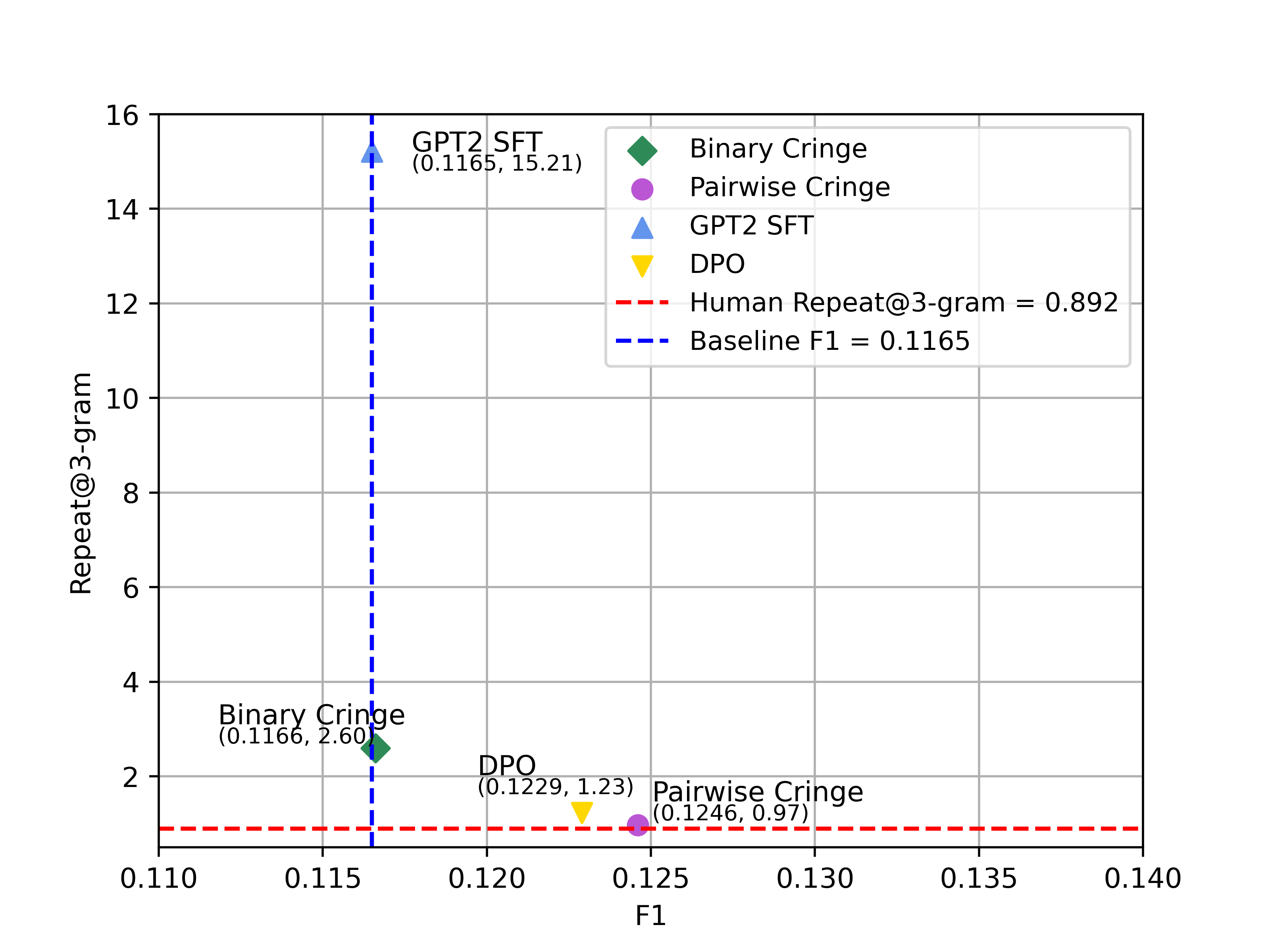}
    \caption{\textbf{Repetition Evaluation}. Test set performance metrics on the repetition mitigation task comparing {{\sc Pairwise Cringe}} with various baselines. {{\sc Pairwise Cringe}} reduces repetitions (Repeat@3-gram) compared to the baseline GPT-2 SFT model generations while improving generation quality (as measured by F1).
    }
    \label{fig:3gramrep}
\end{figure}

\subsection{Reducing Repetitions} \label{sec:repeat}
\paragraph{Training Datasets and Process}
Model-generated completions exhibit sequence-level repetitions, especially with deterministic decoding \cite{holtzman2019curious,unlikelihood_training}. 
{{\sc Pairwise Cringe}} is trained by first supervised fine-tuning GPT2-Medium \cite{radford2019language} on the
BASE data which is a large web-based corpus \cite{lewis2021base} to predict the
next sentence. To construct preference data
to reduce repetitions
one then labels the generations automatically according to whether they contain repeating n-grams or not. 
We generate pairs of outputs from the supervised finetuned GPT2 SFT model using beam blocking decoding (to ensure there are no repetitions) and greedy decoding (which may contain repetitions), and only keep the pair if the generation by greedy decoding contains at least one repeating n-grams (either in the generated sequence itself or a repeat
of the context). The pairwise preferences then use the beam blocked generation as the ``winning'' preferred output, and the greedy decoding with n-grams repeats as the ``losing'' less preferred output. In our experiments we fix $n=3$. We collect in  total  $49,285$ pairs  for this  task. 

We also train {\sc DPO} using the same procedure,  as well as {\sc Binary Cringe} which treats the pairwise preferences as good and bad directly (rather than as a pair, see \autoref{sec:binary_cringe}).

After training, we then generate from
a given model using greedy decoding on the BASE test set, and
the number of repeating n-grams in the generation
(either in the generated sequence itself or a repeat
of the context) is measured, as well as F1 
against the human responses, in order to measure quality. 

\paragraph{Results}
 Results are given in  \autoref{fig:3gramrep}, where the human baseline Repeat@3-gram (by measuring on the responses in the dataset) is 0.892,  whilst the GPT2 SFT model has serious repetition issues for the same contexts, obtaining a  Repeat@3-gram of 15.21 (meaning on average there are 15 n-gram repeats per response), and an F1 of 0.1165.
{\sc Binary Cringe},
{\sc DPO} and
{\sc Pairwise Cringe} all significantly improve over the SFT baseline model in terms of repetitions, with {\sc DPO} and {\sc Pairwise Cringe} providing Repeat@3-gram values close to the human baseline, and {\sc Binary Cringe} slightly trailing. 

In terms of F1, 
{\sc Pairwise Cringe} outperforms {\sc Binary Cringe} significantly, and is slightly higher than DPO as well. 
{\sc DPO} and  {\sc Pairwise Cringe} provide F1 higher than the SFT baseline, whereas {\sc Binary Cringe} does not.

Both {\sc Binary} and {\sc Pairwise Cringe} are run with two iterations, following \autoref{sec:method}.
We can also evaluate the performance of the iteration 1 models. 
Iteration 1 of {\sc Binary Cringe} yields a Repeat@3-
gram value of 1.18 and F1 of 0.1125.
Iteration 1 of {\sc Pairwise Cringe}
yields a Repeat@3-gram value of 1.39 and F1 of 0.1236.
Hence, for both models iteration 1 has worse F1 than the final models.

\subsection{AlpacaFarm Evaluation} \label{sec:alpacafarm}
AlpacaFarm \cite{dubois2023alpacafarm}  is a framework for benchmarking
alignment algorithms that learn to follow user instructions. 
It provides training data in the form of pairwise preferences over responses given to general instruction following tasks.
Additionally, it comes with an automatic evaluation procedure 
using LLMs that was shown to have high agreement with human annotators. 
This framework has been provided in order to evaluate state-of-the-art methods  (PPO, DPO, best-of-n, expert iteration, and more) -- and to compare them to new methods in a controlled environment.
 In the existing results reported, several of those state-of-the-art methods  
 that learn from  preferences are shown 
 to substantially outperform supervised fine-tuning.

\paragraph{Training Datasets and Process}
In line with the training procedure of the benchmark PPO method with human data previously trained in AlpacaFarm \cite{dubois2023alpacafarm}, we leverage the pairwise human preference annotations provided by AlpacaFarm, as well as the identical train sets used in its different RLHF stages:
\begin{itemize}
    \item SFT data: 10k instruction-following demonstrations $(x, y)$ intended for supervised fine-tuning the base LLM to be used in subsequent steps. 
    \item Pairwise-Preference (PREF): 10k instructions with pairwise human feedback data $(x, y^{w}, y^{l})$ collected as part of AlpacaFarm. We note that to compare to standard (binary) Cringe Loss,
    we also convert Pairwise preferences to binary feedback by assigning a {\em positive} label to preferred outputs and  a {\em negative} label to less preferred ones.
    \item Unlabeled: 20k unlabeled instructions $x$ without any responses. We use these for the training iterations of Pairwise Cringe, see \autoref{fig:pco} (bottom).
\end{itemize}

As with the AlpacaFarm baselines we compare against, we start with a Llama-7b model supervised finetuned on the SFT set of the instruction-following demonstrations. 
We then take pairs of human preferences from the PREF set and further finetune the SFT 10k model with different losses for the models we compare, yielding {{\sc DPO}},  {{\sc Binary Cringe}} and {{\sc Pairwise Cringe}}.  

For the {\sc Cringe} models the iterative training is performed using the simple strategy described in \autoref{sec:method}. We start by using the model trained on the PREF set (which we call iteration 1) to generate $k$ responses for each prompt from the Unlabeled set. These are scored using the provided AlpacaFarm reward model ``reward-model-human'' used in AlpacaFarm RLHF training. We then train the second iteration using both the PREF dataset and the newly derived preferences from the Unlabeled set. For both iterations, we start training from the model finetuned on the SFT data. Here we fix $k=4$.

\begin{table}[t]
\caption{AlpacaFarm evaluation results (LLM evaluation), using  human preference data and reward model (where applicable) for training. (*=average of 3 seeds). $^1$PPO with human preferences was  trained by  \citet{dubois2023alpacafarm}; we just evaluated the model.}
\label{tab:alpacafarm_res}
\vskip 0.15in
\begin{center}
\begin{small}
\begin{sc}
\begin{tabular}{lcccr}
\toprule
Method & Win rate (\%) \\
\midrule
{\em Results reported by \citet{dubois2023alpacafarm}}\\
Llama 7B  & 11.3 \\
SFT 10k   & 36.7  \\
SFT 52k   & 39.2 \\
\midrule
{\em Experiments reported in this paper:}\\
Binary {{\sc Cringe}}     & ~47.7* \\
PPO$^1$ & ~48.5* \\
DPO   & ~50.2* \\
Pairwise {{\sc Cringe}}    & ~54.7*  \\
\bottomrule
\end{tabular}
\end{sc}
\end{small}
\end{center}
\vskip -0.1in
\end{table}

\begin{table}[t]
\caption{AlpacaFarm evaluation ablations and further results. (*=average of 3 seeds). $^1$ Result reported from \citet{dubois2023alpacafarm}, uses single seed.}
\label{tab:alpacafarm_ablations}
\vspace{-2mm}
\begin{center}
\begin{small}
\begin{sc}
\begin{tabular}{lcl}
\toprule
Method & Iteration & Win rate (\%) \\
\midrule
{\em Using Human Preferences} \\
Binary {{\sc Cringe}} & 1    & ~~~~45.9* \\
Hard Margin {{\sc Cringe}} & 1 & ~~~~47.8* \\
DPO & 1    & ~~~~50.2* \\
Pairwise {{\sc Cringe}} & 1    & ~~~~52.0* \\ 
\midrule
Binary {{\sc Cringe}} & 2    & ~~~~47.7* \\
Hard Margin {{\sc Cringe}} & 2 & ~~~~49.9* \\
{\sc Iterative DPO} & 2    & ~~~~53.6* \\
Pairwise {{\sc Cringe}} & 2   & ~~~~54.7*  \\
\midrule
\midrule
{\em Using Simulated Preferences} \\
Best-of-n$^1$ & 1            & ~~~~45.0 \\
{\sc Binary Cringe} &  1 & ~~~~45.6*     \\
PPO$^1$ & -              & ~~~~46.8 \\
DPO$^1$ & 1              &  ~~~~46.8\\
{\sc Pairwise Cringe} &  1 & ~~~~50.6*     \\
{\sc Pairwise Cringe} &  2 & ~~~~54.5*     \\
\bottomrule
\end{tabular}
\end{sc}
\end{small}
\end{center}
\vskip -0.1in
\end{table}

\paragraph{Evaluation Dataset} During evaluation, we follow the AlpacaFarm evaluation setup which employs 
LLM-based evaluation, which
selects the superior of two model outputs over 805 prompts, and reports the overall win rates of candidate models against the Davinci-003 model outputs. The 805 instructions in AlpacaFarm evaluation set are sourced from Open Assistant, Anthropic, Vicuna, Koala and self-instruct  evaluations to test models' abilities of following general user instructions. These simulated win rates have shown to have high agreement with human annotations validated by 20k annotations \cite{dubois2023alpacafarm}. 
In our experiments, we report results averaged over 3 seeds.

\paragraph{Main Results}
Our main results are given in \autoref{tab:alpacafarm_res}.
As reported in \citet{dubois2023alpacafarm}, SFT training alone obtains a win rate of 36.7 ({\sc SFT 10k}), or even when training with 52k examples, only improves to a win rate of 39.2 ({\sc SFT 52k}).
These results are outperformed by all the pairwise preference optimization approaches using human preference data. We report the result for the existing AlpacaFarm {\sc PPO} model trained on human preferences, which yields a win rate of 48.5. 
This outperforms {\sc Binary Cringe}, which obtains 47.7.  {\sc DPO} outperforms both of those methods, achieving 50.2. However,
{\sc Pairwise Cringe}  obtains the best performance, with a win rate of
54.7.

\subsubsection{Ablations and further results}

In \autoref{tab:alpacafarm_ablations} we provide additional results.

\paragraph{Pairwise Cringe outperforms Binary Cringe}
First, we find that {\sc Pairwise Cringe} comfortably outperforms {\sc Binary Cringe},
which uses the same pairwise preferences converted to binary feedback, for both training either 1 or 2 iterations.
For example, in iteration 1 of training {\sc Binary Cringe} obtains a win rate
45.9, while {\sc Pairwise Cringe} obtains 52.0.

\paragraph{Soft Margin outperforms Hard Margin Cringe}
Second, for Pairwise Cringe training, we find that a soft margin using a sigmoid gate outperforms a hard margin (win rate  52.0 vs. 47.8) in the first iteration of training, and similarly is better in the second iteration of training as well (win rate 54.7 vs. 49.9). We speculate this is due to the provided gradient available during training in the soft margin case.

\paragraph{Iterations of Cringe training improve performance}
Third, we find that iterations of  {\sc Cringe} improve its win rate.
The first iteration of {\sc Pairwise Cringe} has a win rate of 52.0, while the second iteration has a win rate of 54.7. {\sc Hard Margin Cringe}  and {\sc Binary Cringe} both also benefit from iteration, e.g. an improvement of win rate from 45.9 to 47.7 for {\sc Binary Cringe}, but both still lag behind {\sc Pairwise Cringe}. 

\paragraph{Iterative DPO Improves over DPO}
DPO can also benefit from iteration, which is not the prescribed approach in the original paper. We find performing a second iteration of DPO in the same manner as we perform for our {\sc Cringe} results  (see \autoref{fig:pco}), which we call {\sc Iterative DPO}, results in an improved win rate from 50.2 to 53.6. However, this is still lower than the performance of iteration 2 of {\sc Pairwise Cringe} with 54.7.

\paragraph{Pairwise Cringe performs well on both Human and Simulated Preferences}
While we used human preferences supplied by AlpacaFarm for the experiments so far reported, the original paper also used simulated preferences constructed by LLMs, and reports results for various models with those as well.
Results are shown in \autoref{tab:alpacafarm_ablations}, bottom 5 rows, reporting the numbers from \citet{dubois2023alpacafarm}  for PPO, DPO and Best-of-N.
 We first trained a single iteration of {\sc Binary Cringe} and 
{\sc Pairwise Cringe} in this setting.
{\sc Binary Cringe} obtains a win rate of 45.6, lagging just behind DPO and  PPO (both with 46.8) and slightly ahead of Best-of-N (45.0).
{\sc Pairwise Cringe} (first iteration) provides strong performance, with a win rate of 50.6 -- superior to all other methods tested. Training {\sc Pairwise Cringe} for a second iteration then improves this further, to a win rate of 54.5.

\paragraph{Impact of Hyperparameters}

The hyperparameters used in the experiments are given in \autoref{sec:hypers}.  Common to both standard Cringe and Pairwise Cringe, we find the parameter $\alpha$ is best as being relatively small, in the  0.005-0.01 range.  Like the $\beta$ parameter in DPO, we find the parameter $\tau$ that scales the loss is important to be at the right magnitude, 1-10 in our experiments. The parameter $b$ on the other hand tends to be somewhat less important, but can still give gains from being nonzero. For example, using $b=0$  in iteration one for {\sc Pairwise Cringe Loss} gives a win rate of 50.9, compared to 52.0 for $b=-10$.

\section{Related Work}

Classical large language model  learning 
involves only positive training examples, i.e. modeling the language provided \cite{mikolov2010recurrent,sutskever2014sequence,radford2019language}.
However, if the sequence data distribution is closer to the intended usage then results improve. This motivates pre-training followed by fine-tuning settings where the fine-tune data is also positive examples, but closer to the downstream domain of interest, e.g. dialogue or other tasks \cite{roller2020recipes,gururangan2020don,zhou2023lima}.

Positive example sequences alone, however, 
do not take into account  information about what a model {\em should not do}, which can be captured, amongst other methods, from human feedback.
Human feedback is collected via a user interface (UI), where the type of UI dictates the format of the feedback. 
For example, clicking a thumbs up or down button given a model response \cite{shuster2022blenderbot}
 provides binary feedback data, i.e. good or bad. 
More commonly collected for instruction tuning tasks however, are pairwise preferences indicating response A is preferred to response B \cite{stiennon2020learning,ouyang2022training}. 
The type of data collected dictates the kind of training algorithm that
is then used.

Learning from ranked preferences can be traced back throughout machine learning
history. For example, in the Support Vector Machine (SVM) era, \citet{herbrich1998supervised}  developed
 techniques for learning from ordered preference data using a pairwise margin-based approach. A number of works were developed using related ranking techniques for user preference data, e.g. from web clicks for information retrieval \cite{chapelle2010efficient,chen2008adapting,cao2006adapting}. While many SVM approaches use a hard-margin (Hinge loss), others explored the use of a soft margin, e.g. a sigmoid-type loss as well \cite{perez2000support}. 
 

More recent work, in the deep learning and large language modeling eras,  has focused on viewing preference learning in a reinforcement learning setting.
Typically, a reward model is trained from preference data, and then methods such as
 Proximal Policy Optimization (PPO) \cite{schulman2017proximal}  are applied to fine-tune the language model. Several released models have followed this recipe 
 \cite{ouyang2022training,touvron2023llama}.
 Since then, several other competing approaches have been proposed in particular Direct Preference Optimization (DPO) \cite{rafailov2023direct} which does not require a separate reward model in the loop. 
 Recent models have also been built using DPO \cite{tunstall2023zephyr,mixtralblogpost2023}.
Other models using hard-margin based pairwise approaches have been proposed,
e.g. SliC \cite{zhao2023slic}, CLICK \cite{zheng2023click} and RRHF \cite{yuan2023rrhf} -- although
there is some evidence that the hard margin approach is inferior to DPO
\cite{xu2023contrastive}.
A number of papers have also studied the best way to construct preference pairs,
where better methods 
can result in much improved win rates \cite{yang2023rlcd,zheng2023click,liu2023statistical,xu2023contrastive}.

 Separately, for binary feedback rather than pairwise preferences, several methods have been proposed that aim to train language models using only positive and negative (good and bad) examples.
 The unlikelihood training method \cite{unlikelihood_training,li2019don}, which lowers the probability of generating negative examples, was shown to decrease repetition, copying, and other generation flaws.
 The Cringe Loss \cite{adolphs2022cringe}, itself a generalization of  \citet{simple_contrastive_loss}, was shown to outperform unlikelihood and several other approaches.  For that reason, we chose to study generalizing the Cringe Loss to the pairwise preference case.

 Cringe loss \citep{adolphs2022cringe} first showed that iterative alignment methods can work well, and subsequently other methods have been explored as well, such as
 ReST \citep{gulcehre2023reinforced}, while the concurrent work of \citet{xiong2023gibbs} also studies such approaches for DPO from a more theoretical perspective.

\section{Conclusion}

We introduced the Pairwise Cringe Loss for training language models with pairwise preferences. 
We showed that this approach outperforms its binary feedback counterpart the Cringe Loss, and importantly also outperforms
competing state-of-the-art preference optimization algorithms on the AlpacaFarm benchmark such as PPO and DPO.  
We also showed that the iterative version of our method and Iterative DPO lead to better performance.
Our method is efficient and simple to implement and we expect it
can be applied to a wide range of problems.
We note that our approach can also be used naturally with a mixture of both binary feedback and pairwise preferences if they are available by simply using both versions of the loss (binary Cringe, and Pairwise Cringe) at the same time for the two types of data, making it a versatile choice for end user applications.



\bibliography{example_paper}
\bibliographystyle{icml2024}

\newpage
\appendix
\onecolumn
\section{Appendix}
\label{sec:appendix}


\subsection{Algorithm Details} \label{sec:code}
\hfill%
\begin{minipage}{.98\textwidth}

\begin{lstlisting}[language=Python, caption=Python code for the Cringe Loss., label=python_code]
class CringeLoss(CrossEntropyLoss):
    def __init__(self, alpha=1.0, k=1, **kwargs):
        super().__init__(**kwargs)
        self.alpha = alpha
        self.k = k
    
    def __call__(self, x, y, classifier_labels, **kwargs):
        # Compute the CrossEntropy loss for the positive labels and mask
        # with classifier labels to not train with negative feedback (0)
        ce_loss = super().__call__(x.flatten(0, 1), y.flatten(0, 1), **kwargs)
        cr_loss = self._compute_cringe_loss(x, y, self.k)
        notnull = y.ne(self.ignore_index)
        
        # Remove loss from ignore index.
        ce_loss *= (classifier_labels * notnull).flatten(0, 1)
        cr_loss *= (torch.abs(classifier_labels - 1) * notnull).flatten(0, 1)

        # Compute final loss.
        loss = ce_loss + self.alpha * cr_loss
        return loss, ce_loss, cr_loss
    
    @staticmethod
    def _compute_contrastive_loss(self, x, y, k, **kwargs)
        # compute the contrastive loss part for the negative labels
        # first, get the positives as the top predictions != target
        preds = torch.topk(x, k=k + 1, axis=-1)
        topk_has_tgt = torch.eq(
            preds.indices,
            y.unsqueeze(-1).repeat(1, 1, k + 1),
        )
        topk_logits = preds.values - topk_has_tgt.long() * 1e10

        # if the positive is not in the first k predictions, mask out
        # the final (k+1)'s logit
        topk_logits[:, :, -1] -= (
            torch.logical_not(torch.any(topk_has_tgt, dim=-1)).long()
        ) * 1e10

        # Sample from the categorical distribution of the top-k predictions
        # (with the label masked out).
        preds_dist = Categorical(logits=topk_logits)
        idx_sample = preds_dist.sample()
        sample_preds_values = preds.values.gather(
            dim=-1, index=idx_sample.unsqueeze(-1)
        ).squeeze(-1)

        # Concatenate the logits of the preds with the negative label's logits.
        x_negative_target = x.gather(dim=-1, index=y.unsqueeze(-1)).squeeze(-1)
        x_cr = torch.concat(
            [sample_preds_values.unsqueeze(-1), x_negative_target.unsqueeze(-1)], -1
        )
        # Create the y's for the x_cr (the correct label is always index 0).
        y_cr = torch.zeros_like(y).to(x_cr.device)
        
        # Compute the Cringe Loss as cross entropy loss between x_cr, y_cr
        return F.cross_entropy(x_cr.flatten(0, 1), y_cr.flatten(0, 1), reduction="none")
        
        
\end{lstlisting}
\end{minipage}


\begin{minipage}{.98\textwidth}
\begin{lstlisting}[language=Python, caption=Python code for the Pairwise Cringe Loss., label=python_code]
class PairwiseCringeLoss(CrossEntropyLoss):
    def __init__(self, alpha=1.0, k=1, b=0, tau=1.0, **kwargs):
        super().__init__(**kwargs)
        self.alpha = alpha
        self.k = k
        self.b = b
        self.tau = tau
        assert tau > 0

    def _get_logprob(self, x, y, mask):
        # x: bsz * seqlen * vocab
        # y: bsz * seqlen
        # mask: bsz * seqlen
        token_logps = torch.gather(
            x.log_softmax(-1),
            dim=2,
            index=y.unsqueeze(2),
        ).squeeze(2)
        return (token_logps * mask).sum(dim=-1) / (mask.long().sum(dim=-1) + 1e-10)
    
    def __call__(self, x, y_w, y_l, **kwargs):
        # Compute the CrossEntropy loss for the positive labels
        ce_loss = super().__call__(x.flatten(0, 1), y_w.flatten(0, 1), **kwargs)
        cr_loss = CringeLoss._compute_cringe_loss(x, y_l, self.k)
        mask_l = y_l.ne(self.ignore_index)
        mask_w = y_w.ne(self.ignore_index)
        cr_loss *= mask_l.flatten(0, 1)
        ce_loss *= mask_w.flatten(0, 1)

        # Compute pairwise margin and gate multiplier
        margin = self._get_logprob(x, y_w, mask_w) - self._get_logprob(x, y_l, mask_l)
        sigmoid_gate_multiplier = torch.sigmoid((-margin + self.b)/self.tau)
        sigmoid_gate_multiplier = sigmoid_gate_multiplier.unsqueeze(1)
        
        # Compute final loss
        loss = sigmoid_gate_multiplier * (ce_loss + self.alpha * cr_loss)
        
        return loss, ce_loss, cr_loss
        
\end{lstlisting}

\begin{lstlisting}[language=Python, caption=Python code for the Hard Margin Cringe Loss., label=python_code]
class HardMarginCringeLoss(PairwiseCringeLoss):
    def __call__(self, x, y_w, y_l, **kwargs):
        # Compute the CrossEntropy loss for the positive labels
        ce_loss = super().__call__(x.flatten(0, 1), y_w.flatten(0, 1), **kwargs)
        cr_loss = CringeLoss._compute_cringe_loss(x, y_l, self.k)
        mask_l = y_l.ne(self.ignore_index)
        mask_w = y_w.ne(self.ignore_index)
        cr_loss *= mask_l.flatten(0, 1)
        ce_loss *= mask_w.flatten(0, 1)

        # Compute pairwise margin
        margin = self._get_logprob(x, y_w, mask_w) - self._get_logprob(x, y_l, mask_l)
        margin_based_multiplier = (margin <= self.b).long()
        margin_based_multiplier = margin_based_multiplier.unsqueeze(1)
        
        # Compute final loss
        loss = margin_based_multiplier * (ce_loss + self.alpha * cr_loss)
        
        return loss, ce_loss, cr_loss
        
\end{lstlisting}
    
\end{minipage}

\section{Model Hyperparameters}
\label{sec:hypers}

All the fine-tuned models are trained with a maximum of sixteen 80GB GPUs (NVIDIA A100), optimized with AdamW using $\beta_1 = 0.9$, $\beta_2 = 0.95$, $\epsilon = 1e-08$. Models are trained up to 1000 updates with batch size up to 512. The typical fine-tuning time for a standard decoder-only transformer is less than 3 hrs. 

For all cringe experiments, we fix $topk=5$. For Binary {{\sc Cringe}} on human preferences, the hyperparameters are $
\alpha=0.005$ for iteration 1, and $0.01$ for iteration 2. For Hard Margin {{\sc Cringe}} on human preferences, the  hyperparameters are $\alpha=0.005, b=-10$ for iteration 1 and $\alpha=0.01, b=10$ for iteration 2. For Pairwise {{\sc Cringe}} on human preferences, the Cringe hyperparameters are $\alpha=0.01, b=-10, \tau=10$ for iteration 1 and $\alpha=0.005, b=-10, \tau=1$ for iteration 2. For Pairwise {{\sc Cringe}} on simulated preferences, the Cringe hyperparameters are $\alpha=0.005, b=0, \tau=10$ for iteration 1 and $\alpha=0.01, b=-10, \tau=1$ for iteration 2.

At inference time, we use the same decoding parameters as in AlpacaFarm and sample with temp=0.7,  and set the maximum number of tokens to be 300.
\end{document}